\documentclass{INTERSPEECH2023}
\usepackage{makecell}
\usepackage{multirow, stfloats}
\usepackage{color,soul}
\usepackage{hyperref}
\usepackage{multirow, stfloats}
\usepackage{booktabs}
\usepackage{amssymb}
\usepackage{bbding}
\UseRawInputEncoding

\interspeechcameraready

\title{Assessing Phrase Break of ESL Speech with Pre-trained Language Models and Large Language Models}
\name{Zhiyi Wang$^{1,2*}$, Shaoguang Mao$^2$, Wenshan Wu$^2$, Yan Xia$^2$, Yan Deng$^2$, Jonathan Tien$^2$ \thanks{*Work performed as an intern in Microsoft Research Asia}}
\address{Microsoft Research Asia}
\address{
  $^1$Peking University, Beijing, China\\
  $^2$Microsoft Research Asia, Beijing, China}
\email{wangzhy@stu.pku.edu.cn, shaoguang.mao@microsoft.com, wenshan.wu@microsoft.com, yanxia@microsoft.com, yaden@microsoft.com, jtien@microsoft.com}

\begin{document}

\maketitle
 
\begin{abstract}
This work introduces approaches to assessing phrase breaks in ESL learners' speech using pre-trained language models (PLMs) and large language models (LLMs). There are two tasks: overall assessment of phrase break for a speech clip and fine-grained assessment of every possible phrase break position. To leverage NLP models, speech input is first force-aligned with texts, and then pre-processed into a token sequence, including words and phrase break information. To utilize PLMs, we propose a pre-training and fine-tuning pipeline with the processed tokens. This process includes pre-training with a replaced break token detection module and fine-tuning with text classification and sequence labeling. To employ LLMs, we design prompts for ChatGPT. The experiments show that with the PLMs, the dependence on labeled training data has been greatly reduced, and the performance has improved. Meanwhile, we verify that ChatGPT, a renowned LLM, has potential for further advancement in this area.
\end{abstract}

\section{Introduction}
\label{sec:intro}
Proper phrase break is crucial to oral performance \cite{fach1999comparison} and is always a challenge for English as a Second Language (ESL) learners. There has been considerable research in computer-aided language learning (CALL) \cite{mao2019nn, lin2021improving, mao2022universal, hu2015improved}. As for the phrase break assessment, there are two main categories: 1) break feature extraction and modeling \cite{fu2022using, sabu2018automatic}. 
2) modeling against reference speech \cite{xiao2017proficiency, proencca2019teaching}.
For example, a method was proposed to evaluate break by computing similarity between the assessed speech with utterances from native speakers or Text-to-Speech (TTS) system \cite{xiao2017proficiency}.

\begin{figure}[t]
\centering
\includegraphics[scale=0.5]{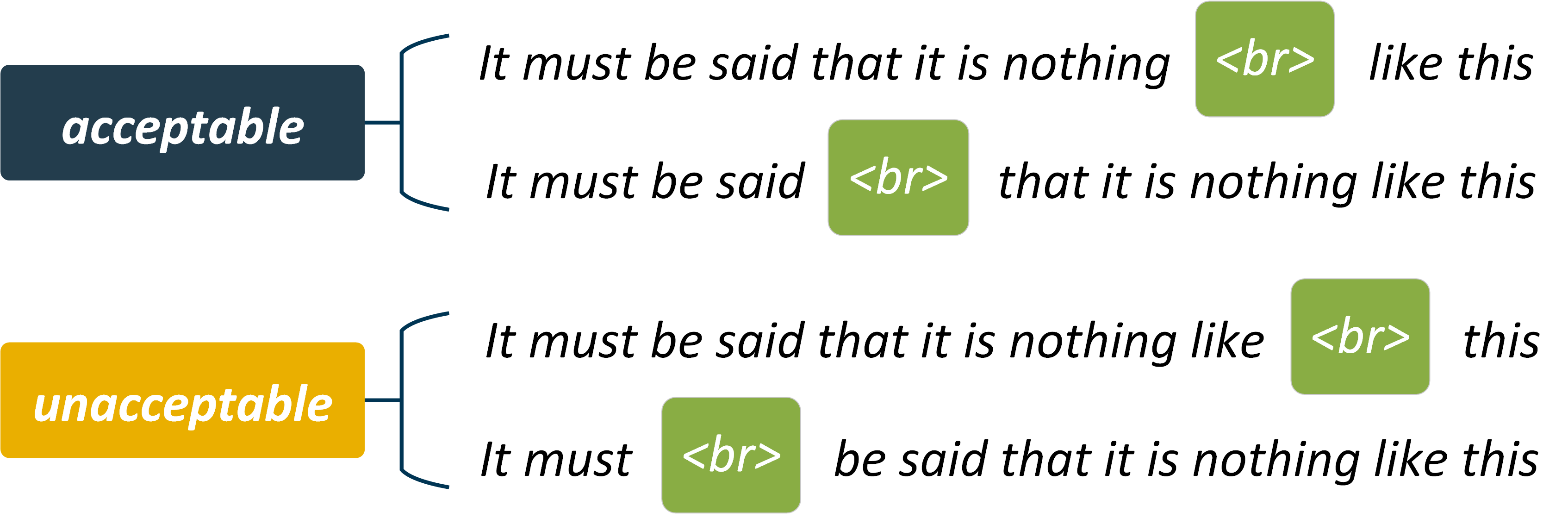}
\caption{Example of diverse phrase break patterns(${<}$br${>}$ represents a phrase break)}\label{fig:break patterns}
\end{figure}

Although modeling against reference speech \cite{xiao2017proficiency, proencca2019teaching} is an effective approach to assessing speech performance, some errors unavoidably occur when it comes to handling diverse phrase break cases. As shown in Figure~\ref{fig:break patterns}, the correct phrase break patterns for the same text are various, and thus it is not to say that the phrasing is incorrect if it is different with template audios. The previous work fails to consider this fact. Instead, they model the phrase break like a fixed pattern prediction. 
Meanwhile, a large scale of high-quality human-labeled data is required for traditional methods. However, the subjective labeling is costly and the labeling consistency is hard to satisfy \cite{mao2019nn, zhang2021speechocean762, meng2010development}. How to construct robust models with small datasets is still under research.

Phrase break prediction is a traditional task in the TTS area \cite{futamata2021phrase, kunevsova2022detection, liu2020exploiting, rendel2016using}. In Futamata's work \cite{futamata2021phrase}, 
a phrase break prediction method is proposed that combines implicit features extracted from BERT \cite{devlin2018bert} and explicit features extracted from Bidirectional Long Short-Term Memory (Bi-LSTM) with linguistic features. The goals of phrase break prediction and phrase break assessment are different, and the second one being much harder considering the diverse break facts. We can refer to the idea that the break information can be inferred from input text, and leverage the power of rising pre-trained language models (PLMs) \cite{liu2019roberta, dong2019unified} and large language models (LLMs) \cite{brown2020language}.




\begin{figure*}[ht]
\centering
\includegraphics[scale=4]{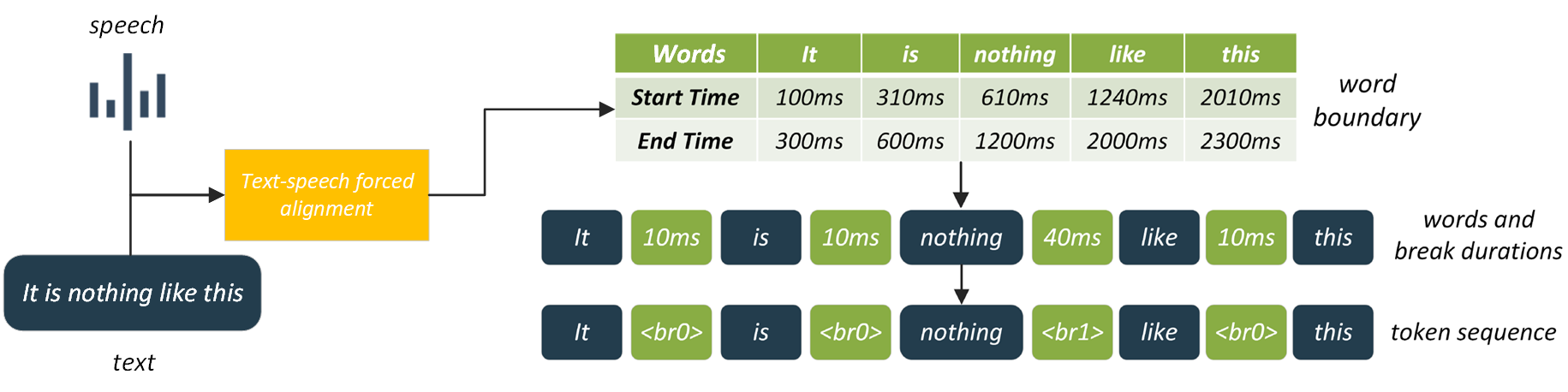}
\caption{Overview of the data pre-processing. The speech-text forced alignment tool recognizes word boundaries and the duration between adjacent words. Then, the token sequence is obtained by converting the duration to break tokens with the mapping in Table~\ref{tab:break_type} .}\label{fig:tokenization}
\end{figure*}

This paper presents approaches to assessing phrase break with PLMs and LLMs. In particular, there are two sub-tasks: assessment of phrase break for a speech and fine-grained assessment of each break position. To adopt those NLP models, each speech is processed into a token sequence with text-speech forced alignment \cite{mathad2021impact, moreno1998recursive, moreno2009factor}, referencing Figure~\ref{fig:tokenization}. The token sequence consists of words and associated phrase break tokens (break duration information for each between-words interval).

To adopt the pre-training models, a self-supervised replaced break token detection strategy is proposed. Each break token from the original sample has 15\% chance of being replaced by other break tokens. Then, a discriminator is trained with augmented data riding on BERT \cite{devlin2018bert} to identify whether the token sequence is edited. In the fine-tuning stage, the overall assessment and fine-grained assessment are fine-tuned with text classification and token classification, respectively. Additionally, by providing suitable prompts, LLMs can perform well on many NLP tasks and adapt for specific use-cases with just a few task examples \cite{brown2020language, chen2021evaluating, huang2022towards}. Therefore, we design prompts and investigate the zero-shot and few-shot learning \cite{ravi2017optimization, sung2018learning, wang2020generalizing} setups with ChatGPT \cite{OpenAI2021}.

The main contributions are: first, this is pioneering work to explore the use of PLMs and LLMs for speech assessment. The experimental results demonstrate the possibility of using language models to perform speech evaluation in specific tasks. Second, this work takes diverse phrasing patterns into consideration to construct a more precise assessment.



\section{Data Pre-processing}
\label{sec:format}
\subsection{Task definition}
\label{subsec:task-def}
We use two tasks to demonstrate how PLMs and LLMs can be applied into assessing phrase break.

One is predicting a rank $r$ for a test speech to indicate its overall performance on phrase break. The other one is that given a speech $S$, consisting $n$ words, predict a rank $r_i$ for each interval $b_i$ between two words on whether the phrase break is appropriate, including whether an existing break is appropriate and whether an expected break is missed.


\subsection{Pre-processing}
\label{sec:pre-processing}
To leverage the power of PLMs and LLMs, the speech clips are first converted to a token sequence with speech-text forced alignment. 

As shown in Figure~\ref{fig:tokenization}, speech-text forced alignment is used to recognize the phrase break and duration between every pair of adjacent words $w_i$ and $w_{i+1}$. Based upon the statistical information and linguists' assessment, the phrase breaks are categorized into four types, as shown in Table~\ref{tab:break_type}. A speech utterance is then tokenized into a token sequence $T:\left\{w_0,b_0,w_1,...,w_i,b_i,w_{i+1},...,w_n\right\}$, including words and phrase break tokens. 

\section{Approach}
\subsection{Pre-trained Language Models for Break Assessment}
\label{sec:pre-training}
\begin{figure*}[ht]
\centering
\includegraphics[scale=0.5]{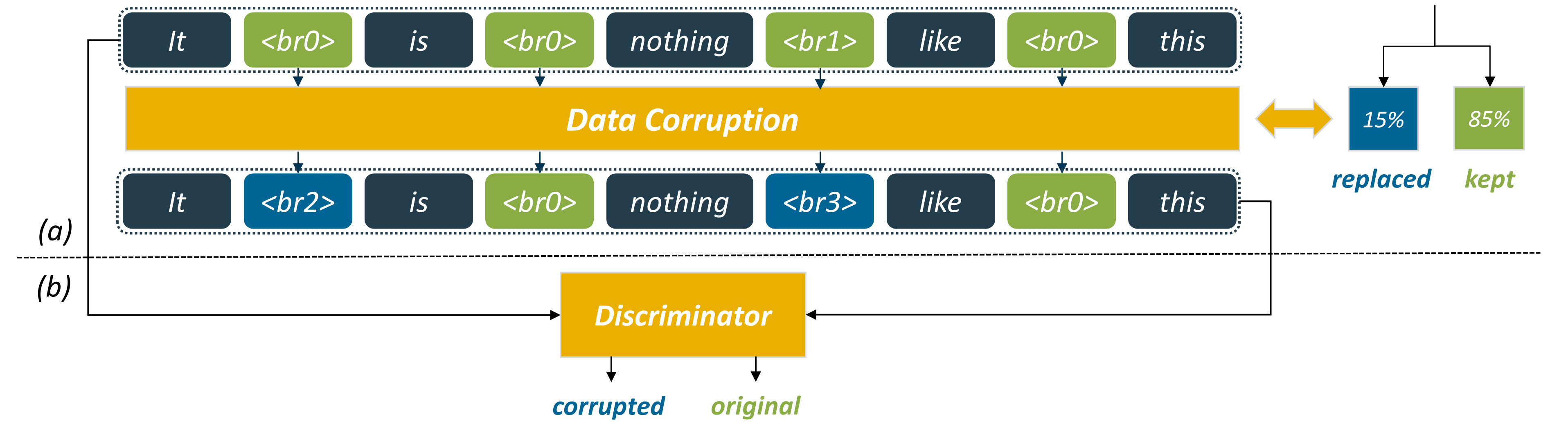}
\caption{An overview of the replaced break token detection pre-training process. Part (a) describes an example of data corruption where each break token in the original token sequence has 15\% chance of being replaced with other tokens. Part (b) is the pre-training stage where a discriminator is trained to distinguish original or corrupted data.}\label{fig:data_corruption}
\end{figure*}

\textbf{Replaced Break Token Detection}
We introduce a pre-training approach named replaced break token detection. As shown in Figure~\ref{fig:data_corruption}, speech recordings by native speakers from TTS corpus are collected as original samples because TTS recordings have good performance in phrase break. Then, each sample is randomly corrupted several times with the strategy that each break token has 15\% chance to be replaced with other kinds of break tokens. 15\% is a hyper-parameter settled by pre-experiments. The proportions of different types of breaks after the random corruptions with a 15\% change and the real speech by non-native speakers are very similar.

\begin{table}
\centering
\caption{The definition of break tokens}
\label{tab:break_type}
\begin{tabular}{lll}
\hline
\textbf{Type} & \textbf{Duration} & \textbf{Comment}\\
\hline
{br0} & {(0, 10ms]} & {No break} \\
{br1} & {(10ms, 50ms]} & {Slight / Optional break} \\
{br2} & {(50ms, 200ms]} & {Break} \\
{br3} & {(200ms, $+ \infty$)} & {Long break} \\
\hline
\end{tabular}
\end{table}

The pre-training is from BERT as it is trained on a large scale of texts and learns contextual relations between words (or sub-words) \cite{devlin2018bert}. 
A discriminator is trained with cross-entropy loss on the augmented data to predict whether the input sequence has been corrupted. The trained model is called Break-BERT for convenience.

\textbf{Downstream Tasks}
The overall assessment task is treated as a sequence classification task.
The model predicts a rank $r$ for a token sequence, $r \in R$. 
The model consists of the head of the pre-trained model and a classifier on top, and is trained with cross-entropy loss.

The fine-grained assessment is treated as a sequence labeling task. An $r_i \in R$ is expected to be assigned to $b_i$. There is a token classification head on top of the hidden-state output from pre-trained model. It is also trained with a cross-entropy loss function. 

\subsection{Large Language Models for Break Assessment}
We investigate the potential of ChatGPT for phrase break assessment in zero-shot and few-shot scenarios.

\textbf{Prompts} By taking into account the crucial impact that prompting has on output from generative models, we clarify our problem according to section \ref{subsec:task-def}, input scoring rubric that annotators adopt, and then standardize input and output formats for our task. 

The input is formatted as $T$ in section \ref{sec:pre-processing}. The output is formatted as rank $r, r\in R$ and inappropriate break position set $P:\{p_1,...,p_j,...p_n\}, p_j=w_i b_i w_{i+1}$ represents that the phrase break $b_i$ between word $w_i$ and word $w_{i+1}$ is inappropriate.

\begin{figure}[h]
\centering
\includegraphics[scale=0.28]{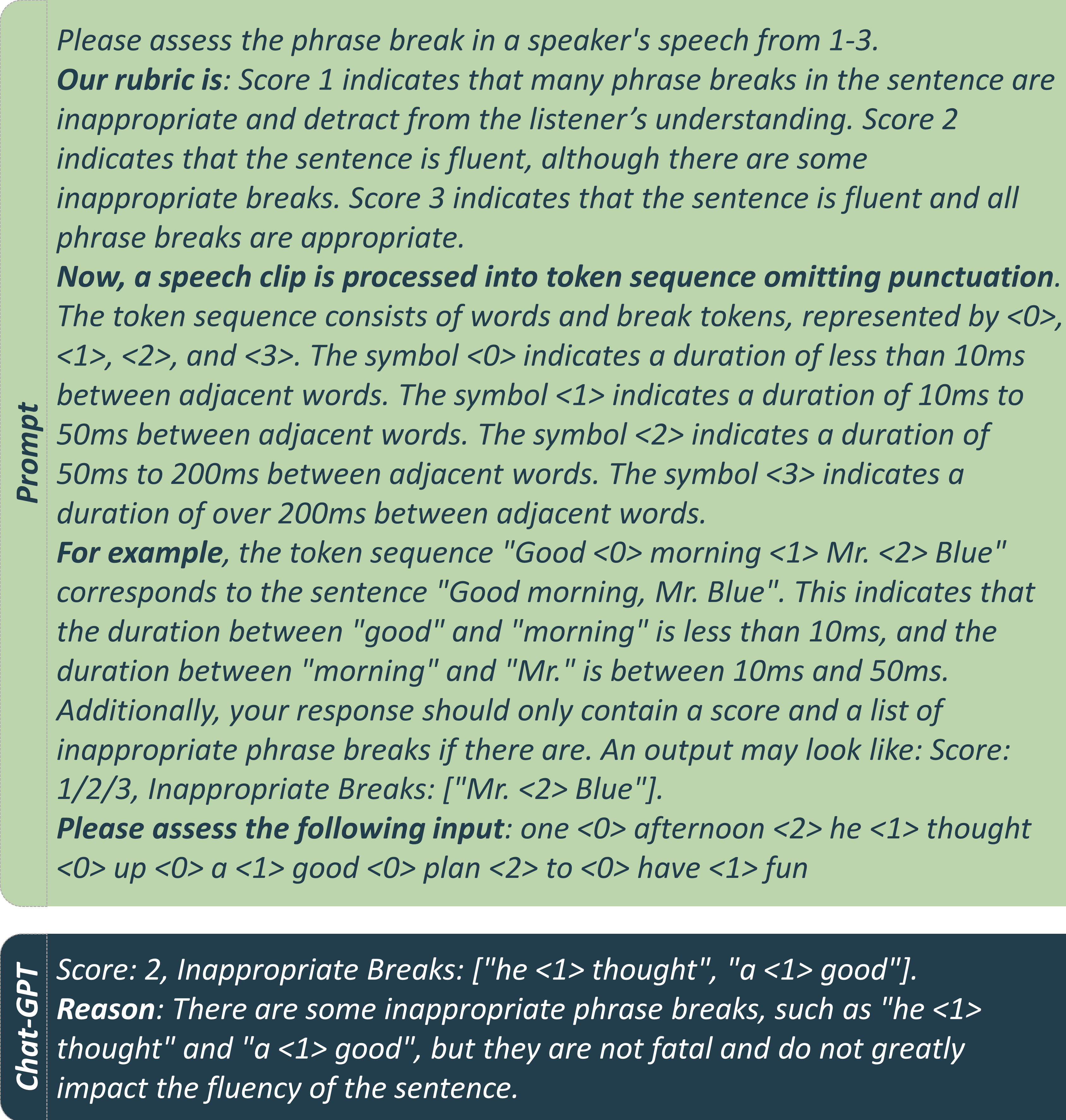}
\caption{An example of prompt and response from ChatGPT.}\label{fig:prompt-gpt}
\end{figure}

\textbf{Zero Shot or Few Shot Learning}
We try zero-shot and few-shot learning to explore the potential of LLMs in speech phrase break assessment. For the zero-shot scenario, ChatGPT responses without specifically training on any data. A zero-shot example of question prompt and response is shown in Figure~\ref{fig:prompt-gpt}. For the few-shot learning, a few examples are provided as context information to enable the model be adapted for annotated cases. Details for case selection in few-shot learning will be discussed in the next section.

\section{Experiments}
\label{sec:pagestyle}

\subsection{Corpora}
\label{subsec:corpora}
We collected 800 audio samples from different Chinese ESL learners. Then, two linguists are invited to assess them with overall performance on phrase break and each individual phrase break ranging from 1 (Poor), 2 (Fair), and 3 (Great). If two experts' opinions are inconsistent, an extra linguist will intervene and do the final scoring. The statistics of collected corpus are listed in Table~\ref{tab:finetune_dataset}. All data is publicly available for research on \url{https://github.com/Chris0King/phrasing-break-assessment}.

\begin{table}[h]\centering
\caption{Statistics of downstream datasets.}
\label{tab:finetune_dataset}
\begin{tabular}{c|ccc|c}
\toprule
\textbf{Dataset} &\textbf{Poor} & \textbf{Fair} & \textbf{Great} & \textbf{Total}\\
\hline
{Overall} & {21} & {136} & {643} & {800}\\
\hline
{Fine-grained} & {129} & {644} & {10797} & {11570}\\
\bottomrule
\end{tabular}
\end{table}

\subsection{Pre-training Setups}
The data for the downstream task was obtained from recordings of 800 Chinese students during "read-after-me" exercises. Thus, we utilized the TTS corpus, a speech dataset from reading scenarios, for pre-training purposes. The pre-training was performed using the LJ Speech dataset  \cite{ljspeech17}, which is commonly known for its good phrase breaks and diverse break patterns. There are 22.5 hours of speech recordings in the training set and 1 hour in the test set, containing 192k words and 8k words respectively.

In data corruption, the ratio between the corrupted samples and the original samples is 3:1, i.e. for each original sample, three random corrupted samples are augmented. The pre-training begins from BERT$_\textbf{BASE}$, and a simple linear classifier is added on the top. It is trained with a batch size of 64 for 3 epochs over the dataset. The maximum sequence length is set to 128. We used back propagation and Adam optimizer with a learning rate of 1e-4. After the pre-training, the binary classification results of Break-BERT achieve 83.9\% in accuracy and 89.7\% in f-score.

\begin{table*}[t]\centering
\caption{Performance of overall and fine-grained assessment models. '\#' stands for 'Fine-tune' and 'w/' stands for 'with'.}
\label{tab:overall_scoring}
\begin{tabular}{c|c|ccc}
\toprule
\multicolumn{2}{c|}{\multirow{2}*{\textbf{Assessment Model}}}
& \multicolumn{3}{c}{\textbf{Metric avg.(std)}}\\
\cmidrule(r){3-5}
\multicolumn{2}{c|}{~} &{Acc.} & {F-Score(weighted)}& {F-Score(macro)}\\
\hline 
\multirow{6}*{\textbf{Overall}}&{Bi-LSTM} & {80.2(6.4)} & {76.4(9.6)}& {39.2(8.1)}\\
&{Against-TTS} & {54.4(9.9)} & {61.1(7.1)}& {36.3(5.6)} \\
&{\#BERT} & {80.4(6.5)} & {77.9(7.0)}& {40.9(7.1)} \\
&{\#Break-BERT} & \textbf{82.5(5.0)} & \textbf{81.7(5.7)}& \textbf{52.3(10.5)} \\
&{ChatGPT w/ Zero-shot Learning} & {55.6(6.2)} & {61.6(4.9)}& {40.6(3.7)} \\
&{ChatGPT w/ Few-shot Learning} & {65.8(5.8)} & {70.5(4.8)}& {47.3(4.0)} \\
\hline
\multirow{4}*{\textbf{Fine-grained}}&{Bi-LSTM} & {92.5(3.9)} & {90.1(5.6)} & {39.9(3.7)}\\
&{Against-TTS} & {70.9(2.6)} & {78.6(4.0)}& {31.1(1.5)} \\
&{\#BERT} & {91.8(4.1)} & {89.0(5.8)} & {39.5(4.1)}\\
&{\#Break-BERT} & \textbf{92.8(3.1)} & \textbf{91.6(4.0)} & \textbf{44.3(2.5)}\\
\bottomrule
\end{tabular}
\end{table*}


\begin{table*}[h]\centering
\caption{Comparing the performance of \#Break-Bert and ChatGPT on fine-grained assessment, category 1 (Poor) and category 2 (Fair) are mapped to inappropriate break, and category 3 (Great) is mapped to appropriate break.}
\label{tab:chatgpt_fine-grained}
\begin{tabular}{c|c|c|c|c|c|c}
\toprule
~ & \multicolumn{2}{c|}{\textbf{\#Break-BERT}} & \multicolumn{2}{c|}{\textbf{ChatGPT w/ Zero-shot}} &\multicolumn{2}{c}{\textbf{ChatGPT w/ Few-shot}}\\
\hline
{Category} & {Precision} & {Recall} & {Precision} & {Recall} & {Precision} & {Recall}\\
\hline
{Poor and Fair} & {\textbf{60.1\%}} & {33.7\%} & {26.5\%} & {31.4\%}& {26.9\%} & {32.6\%}\\
{Great} & {\textbf{94.8\%}} & {\textbf{98.4\%}} & {94.4\%} & {94.4\%}& {94.5\%} & {93.1\%}\\
\bottomrule
\end{tabular}
\end{table*}


\subsection{Experimental Setup}
\textbf{Baselines}
Bi-LSTM+Linear Layer and Bi-LSTM+CRF (Conditional Random Field) \cite{rei2017semi} are set as baselines for overall and fine-grained assessment, separately. We apply Bi-LSTM as a backbone considering it still works well in a relatively small dataset. The hidden layer size is set to 1024. Meanwhile, a direct fine-tuning with downstream data on BERT is conducted to verify the validity of the proposed pre-training process. The baseline models take the identical token sequences by the BERT tokenizer.

We also adopt the Against-TTS method  \cite{xiao2017proficiency} as a baseline and tag the output break similarity score [0, 0.3), [0.3, 0.7), [0.7, 1.0] as poor, fair, great, separately. The adopted TTS system is from Microsoft Cognitive Service en-US-AriaNeural \footnote{\url{https://learn.microsoft.com/en-us/azure/cognitive-services/}}.

Meanwhile, ChatGPT was asked to assess the phrase break from 1-3 defined in \ref{subsec:corpora} and list all inappropriate breaks. For few-shot learning, we evaluate each example in the test set by randomly selecting four samples from the corresponding training set as context information to maintain the similar distribution. The prompt and few-shot learning strategy are determined through preliminary experiments. All examples are tested on text-chatdavinci-002 (using OpenAI's playground). The temperature is set 0 to ensure a consistent prediction. 

\textbf{Cross-validation} 
We apply five-fold cross-validation to avoid instability of sampling and report the mean and standard deviation of experiments.


\subsection{Results}

Accuracy, weighted f-score and macro f-score are taken as metrics \cite{zhang2021speechocean762}. As shown in Table~\ref{tab:overall_scoring} and Table~\ref{tab:chatgpt_fine-grained}, compared with Bi-LSTM, Against-TTS, fine-tuning on BERT and ChatGPT, the proposed pre-training fine-tuning greatly improves all metrics. The knowledge learned from the pre-training stage efficiently enhances model performance. 
It is worth mentioning that the Against-TTS system performs much worse than the proposed approach and ChatGPT. More discussions are included in the next section.

\section{Discussion}

\subsection{Influence of Pre-training}
The pre-training process takes TTS human recordings as correct samples, where multiple phrase break patterns exist. After a series of random corruptions, the augmented samples are likely to be incorrect in phrasing. After the pre-training on original and constructed incorrect patterns, the discriminator has learned general linguistic patterns and phrase break information through self-supervised learning. 
The experiments verified the assumptions. The proposed model yields better results. The knowledge learned from pre-training benefits downstream tasks.


\subsection{How Diverse Breaks are Handled}
The experimental results verified Against-TTS approach's limits on handling multiple possible phrase breaks.  As shown in Table~\ref{tab:confusing_matric}, there are sharp drops of the recall of category 3 (Great) and the precision of category 1 (Poor), while the precision of category 3 (Great) and the recall of category 1 (Poor) are kept. For a test speech, if it shows a different phrase break pattern with reference audio, it tends to be classified as poor even if it is correct. 
When it shows a similar phrase break pattern to the template, it is highly possible to be a correct phrasing. This explains the high precision, low recall for category 3 (Great), as well as the high recall, low precision of category 1 (Poor). 

\begin{table}[h]\centering
\caption{Performance analysis on different categories. }
\label{tab:confusing_matric}
\begin{tabular}{c|c|c|c|c}
\toprule
~ & \multicolumn{2}{c|}{\textbf{Against-TTS}} & \multicolumn{2}{c}{\textbf{\#Break-BERT}}\\
\hline
{Category} & {Precision} & {Recall} & {Precision} & {Recall}\\
\hline
{Poor} & {4.3\%} & {\textbf{28.6\%}} & {\textbf{50.0\%}} & {14.3\%}\\
{Fair} & {26.7\%} & {46.3\%} & {\textbf{49.2\%}} & {\textbf{47.8\%}}\\
{Great} & {87.3\%} & {57.5\%} & {\textbf{89.7\%}} & {\textbf{92.4\%}}\\
\bottomrule
\end{tabular}
\end{table}

\subsection{LLMs for Break Assessment}


According to experiments, we noticed that the ChatGPT model with zero-shot learning exhibits partial understanding of punctuation breaks. However, it tends to overlook the slight pauses between semantic groups, such as the initiation of a clause or phrase. Despite some improvement with few-shot learning setting, the ChatGPT model still struggles to adequately address the breaks between semantic groups and manifests unstable performance when attempting to rectify incorrect breaks.

While not reaching the state-of-the-art performance, the potential of ChatGPT in phrase break assessment is noteworthy. After few-shot learning, all metrics improve significantly in overall assessment task.
We believe that with further optimization in prompt design, ChatGPT has the potential to demonstrate greater power in speech assessment.


\section{Conclusions}
\label{sec:typestyle}
This work presents new approaches to tackling ESL speech phrase break assessment with pre-trained language models (PLMs) and large language models (LLMs). The introduction of PLMs greatly minimizes the requirements for collecting labeled data, and the proposed self-supervised learning can handle multiple possible phrase break patterns of the same text. Also, we verify that ChatGPT, a classical and renowned LLM, has potential for further advancement in this area.
In the future, leveraging PLMs and LLMs to solve other prosody assessment tasks, like intonation and stress, is well worth researching.




\bibliographystyle{IEEEtran}
\bibliography{mybib}

\begin{thebibliography}{10}
\providecommand{\url}[1]{#1}
\csname url@samestyle\endcsname
\providecommand{\newblock}{\relax}
\providecommand{\bibinfo}[2]{#2}
\providecommand{\BIBentrySTDinterwordspacing}{\spaceskip=0pt\relax}
\providecommand{\BIBentryALTinterwordstretchfactor}{4}
\providecommand{\BIBentryALTinterwordspacing}{\spaceskip=\fontdimen2\font plus
\BIBentryALTinterwordstretchfactor\fontdimen3\font minus
  \fontdimen4\font\relax}
\providecommand{\BIBforeignlanguage}[2]{{%
\expandafter\ifx\csname l@#1\endcsname\relax
\typeout{** WARNING: IEEEtran.bst: No hyphenation pattern has been}%
\typeout{** loaded for the language `#1'. Using the pattern for}%
\typeout{** the default language instead.}%
\else
\language=\csname l@#1\endcsname
\fi
#2}}
\providecommand{\BIBdecl}{\relax}
\BIBdecl

\bibitem{fach1999comparison}
M.~Fach, ``A comparison between syntactic and prosodic phrasing.'' in
  \emph{Eurospeech}, vol.~99.\hskip 1em plus 0.5em minus 0.4em\relax Citeseer,
  1999, pp. 527--530.

\bibitem{mao2019nn}
S.~Mao, Z.~Wu, J.~Jiang, P.~Liu, and F.~Soong, ``Nn-based ordinal regression
  for assessing fluency of esl speech,'' in \emph{Proc. ICASSP}.\hskip 1em plus
  0.5em minus 0.4em\relax IEEE, 2019, pp. 7420--7424.

\bibitem{lin2021improving}
B.~Lin, L.~Wang, H.~Ding, and X.~Feng, ``Improving l2 english rhythm evaluation
  with automatic sentence stress detection,'' in \emph{2021 IEEE Spoken
  Language Technology Workshop (SLT)}.\hskip 1em plus 0.5em minus 0.4em\relax
  IEEE, 2021, pp. 713--719.

\bibitem{mao2022universal}
S.~Mao, F.~Soong, Y.~Xia, and J.~Tien, ``A universal ordinal regression for
  assessing phoneme-level pronunciation,'' in \emph{Proc. ICASSP}.\hskip 1em
  plus 0.5em minus 0.4em\relax IEEE, 2022, pp. 6807--6811.

\bibitem{hu2015improved}
W.~Hu, Y.~Qian, F.~Soong, and Y.~Wang, ``Improved mispronunciation detection
  with deep neural network trained acoustic models and transfer learning based
  logistic regression classifiers,'' \emph{Speech Communication}, vol.~67, pp.
  154--166, 2015.

\bibitem{fu2022using}
K.~Fu, S.~Gao, X.~Tian, W.~Li, Z.~Ma, and A.~Bytedance, ``Using fluency
  representation learned from sequential raw features for improving non-native
  fluency scoring,'' in \emph{Proc. Interspeech}, 2022, pp. 4337--4341.

\bibitem{sabu2018automatic}
K.~Sabu and P.~Rao, ``Automatic assessment of children's oral reading using
  speech recognition and prosody modeling,'' \emph{CSI Transactions on ICT},
  vol.~6, no.~2, pp. 221--225, 2018.

\bibitem{xiao2017proficiency}
Y.~Xiao and F.~Soong, ``Proficiency assessment of esl learner's sentence
  prosody with tts synthesized voice as reference.'' in \emph{INTERSPEECH},
  2017, pp. 1755--1759.

\bibitem{proencca2019teaching}
J.~Proença, G.~Raboshchuk, Ã.~Costa, P.~Lopez-Otero, and X.~Anguera,
  ``Teaching american english pronunciation using a tts service.'' in
  \emph{SLaTE}, 2019, pp. 59--63.

\bibitem{zhang2021speechocean762}
J.~Zhang, Z.~Zhang, Y.~Wang, Z.~Yan, Q.~Song, Y.~Huang, K.~Li, D.~Povey, and
  Y.~Wang, ``speechocean762: An open-source non-native english speech corpus
  for pronunciation assessment,'' \emph{arXiv preprint arXiv:2104.01378}, 2021.

\bibitem{meng2010development}
H.~Meng, W.~Lo, A.~Harrison, P.~Lee, K.~Wong, W.~Leung, and F.~Meng,
  ``Development of automatic speech recognition and synthesis technologies to
  support chinese learners of english: The cuhk experience,'' in \emph{Proc.
  APSIPA ASC}, 2010, pp. 811--820.

\bibitem{futamata2021phrase}
K.~Futamata, B.~Park, R.~Yamamoto, and K.~Tachibana, ``Phrase break prediction
  with bidirectional encoder representations in japanese text-to-speech
  synthesis,'' \emph{arXiv preprint arXiv:2104.12395}, 2021.

\bibitem{kunevsova2022detection}
M.~Kunešová and M.~Řezáčková, ``Detection of prosodic boundaries in
  speech using wav2vec 2.0,'' in \emph{International Conference on Text,
  Speech, and Dialogue}.\hskip 1em plus 0.5em minus 0.4em\relax Springer, 2022,
  pp. 377--388.

\bibitem{liu2020exploiting}
R.~Liu, B.~Sisman, F.~Bao, J.~Yang, G.~Gao, and H.~Li, ``Exploiting
  morphological and phonological features to improve prosodic phrasing for
  mongolian speech synthesis,'' \emph{IEEE/ACM Transactions on Audio, Speech,
  and Language Processing}, vol.~29, pp. 274--285, 2020.

\bibitem{rendel2016using}
A.~Rendel, R.~Fernandez, R.~Hoory, and B.~Ramabhadran, ``Using continuous
  lexical embeddings to improve symbolic-prosody prediction in a text-to-speech
  front-end,'' in \emph{Proc. ICASSP}.\hskip 1em plus 0.5em minus 0.4em\relax
  IEEE, 2016, pp. 5655--5659.

\bibitem{devlin2018bert}
J.~Devlin, M.~Chang, K.~Lee, and K.~Toutanova, ``Bert: Pre-training of deep
  bidirectional transformers for language understanding,'' \emph{arXiv preprint
  arXiv:1810.04805}, 2018.

\bibitem{liu2019roberta}
Y.~Liu, M.~Ott, N.~Goyal, J.~Du, M.~Joshi, D.~Chen, O.~Levy, M.~Lewis,
  L.~Zettlemoyer, and V.~Stoyanov, ``Roberta: A robustly optimized bert
  pretraining approach,'' \emph{arXiv preprint arXiv:1907.11692}, 2019.

\bibitem{dong2019unified}
L.~Dong, N.~Yang, W.~Wang, F.~Wei, Y.~Liu, X.and~Wang, J.~Gao, M.~Zhou, and
  H.~Hon, ``Unified language model pre-training for natural language
  understanding and generation,'' \emph{Advances in Neural Information
  Processing Systems}, vol.~32, 2019.

\bibitem{brown2020language}
T.~Brown, B.~Mann, N.~Ryder, M.~Subbiah, J.~D. Kaplan, P.~Dhariwal,
  A.~Neelakantan, P.~Shyam, G.~Sastry, A.~Askell \emph{et~al.}, ``Language
  models are few-shot learners,'' \emph{Advances in neural information
  processing systems}, vol.~33, pp. 1877--1901, 2020.

\bibitem{mathad2021impact}
V.~Mathad, T.~Mahr, N.~Scherer, K.~Chapman, K.~Hustad, J.~Liss, and V.~Berisha,
  ``The impact of forced-alignment errors on automatic pronunciation
  evaluation.'' in \emph{Interspeech}, 2021, pp. 1922--1926.

\bibitem{moreno1998recursive}
P.~Moreno, C.~Joerg, J.~Van~Thong, and O.~Glickman, ``A recursive algorithm for
  the forced alignment of very long audio segments.'' in \emph{ICSLP}, vol.~98,
  1998, pp. 2711--2714.

\bibitem{moreno2009factor}
P.~Moreno and C.~Alberti, ``A factor automaton approach for the forced
  alignment of long speech recordings,'' in \emph{Proc. ICASSP}.\hskip 1em plus
  0.5em minus 0.4em\relax IEEE, 2009, pp. 4869--4872.

\bibitem{chen2021evaluating}
M.~Chen, J.~Tworek, H.~Jun, Q.~Yuan, H.~P. d.~O. Pinto, J.~Kaplan, H.~Edwards,
  Y.~Burda, N.~Joseph, G.~Brockman \emph{et~al.}, ``Evaluating large language
  models trained on code,'' \emph{arXiv preprint arXiv:2107.03374}, 2021.

\bibitem{huang2022towards}
J.~Huang and K.~C.-C. Chang, ``Towards reasoning in large language models: A
  survey,'' \emph{arXiv preprint arXiv:2212.10403}, 2022.

\bibitem{ravi2017optimization}
S.~Ravi and H.~Larochelle, ``as a model for few-shot learning,'' in
  \emph{International conference on learning representations}, 2017.

\bibitem{sung2018learning}
F.~Sung, Y.~Yang, L.~Zhang, T.~Xiang, P.~H. Torr, and T.~M. Hospedales,
  ``Learning to compare: Relation network for few-shot learning,'' in
  \emph{Proceedings of the IEEE conference on computer vision and pattern
  recognition}, 2018, pp. 1199--1208.

\bibitem{wang2020generalizing}
Y.~Wang, Q.~Yao, J.~T. Kwok, and L.~M. Ni, ``Generalizing from a few examples:
  A survey on few-shot learning,'' \emph{ACM computing surveys (csur)},
  vol.~53, no.~3, pp. 1--34, 2020.

\bibitem{OpenAI2021}
OpenAI, ``Chatgpt,'' \url{https://chat.openai.com/}, 2022.

\bibitem{ljspeech17}
K.~Ito and L.~Johnson, ``The lj speech dataset,''
  \url{https://keithito.com/LJ-Speech-Dataset/}, 2017.

\bibitem{rei2017semi}
M.~Rei, ``Semi-supervised multitask learning for sequence labeling,''
  \emph{arXiv preprint arXiv:1704.07156}, 2017.

\end{thebibliography}

\end{document}